# Multi-Reference Path Tracking Control for an Agricultural Tractor with Nonlinear Model Predictive Control

**Marcel Moll** * **Timo Oksanen** *

* *Chair of Agrimechatronics, Munich Institute of Robotics and Machine Intelligence (MIRMI), Technical University of Munich, Germany (e-mail: {marcel.moll; timo.oksanen}@tum.de)*

**Abstract:** Guiding a tractor along a predefined reference path is a key component of precision agriculture. This study develops a path tracking controller based on Nonlinear Model Predictive Control, which incorporates multiple segments of a piecewise-linear reference path directly into the objective function. In addition, methods for selecting viable reference segments from the full path are presented. The control system is evaluated during a field test with a tractor controlled via the Tractor Implement Management steering interface. The NMPC solver converged on average after 3.45 ms and tracked the curved reference path with a mean absolute cross-track error of 6.1 cm.

## 1. INTRODUCTION

The main task of a tractor guidance system is to follow a predefined path with minimal deviation. While this task was performed manually in the era when horses were more common than tractors, the demanding work had already been eased well before the advent of electronics in agriculture. As noted in Heraud and Lange (2009), a wide range of mechanical aids had been invented and described in patents.

With the public availability of Global Navigation Satellite System (GNSS) receivers capable of achieving centimeter-level accuracy through Real-Time Kinematic (RTK)-based ambiguity resolution methods, automatic guidance systems have become an integral part of precision agriculture.

### *1.1 Current State of Research*

An essential component of an electronic guidance system is the underlying control algorithm. Since a standard tractor has similar kinematic properties as most on-road vehicles, a variety of control methodologies have been proposed in the literature and applied to agricultural guidance systems. This work focuses on systems employing Model Predictive Control (MPC), which compute control outputs in real time using an online optimization procedure with a limited prediction horizon.

Time-based trajectory tracking for agricultural tractors is demonstrated in Kraus et al. (2013) using an Nonlinear Model Predictive Control (NMPC) combined with an Moving Horizon Estimation. The time-parameterized reference trajectory steers the vehicle to specified states at predefined times, which is according to Kayacan et al. (2018) necessary for multi-machine applications. Building on a similar type of reference, Kayacan et al. (2018) compares an input-linearized MPC with a nonlinear NMPC for a tractor towing a steerable trailer.

The project is supported by funds of the Federal Ministry of Agriculture, Food and Regional Identity (BMLEH) based on a decision of the Parliament of the Federal Republic of Germany via the Federal Office for Agriculture and Food (BLE) under the strategy for digitalisation in agriculture.

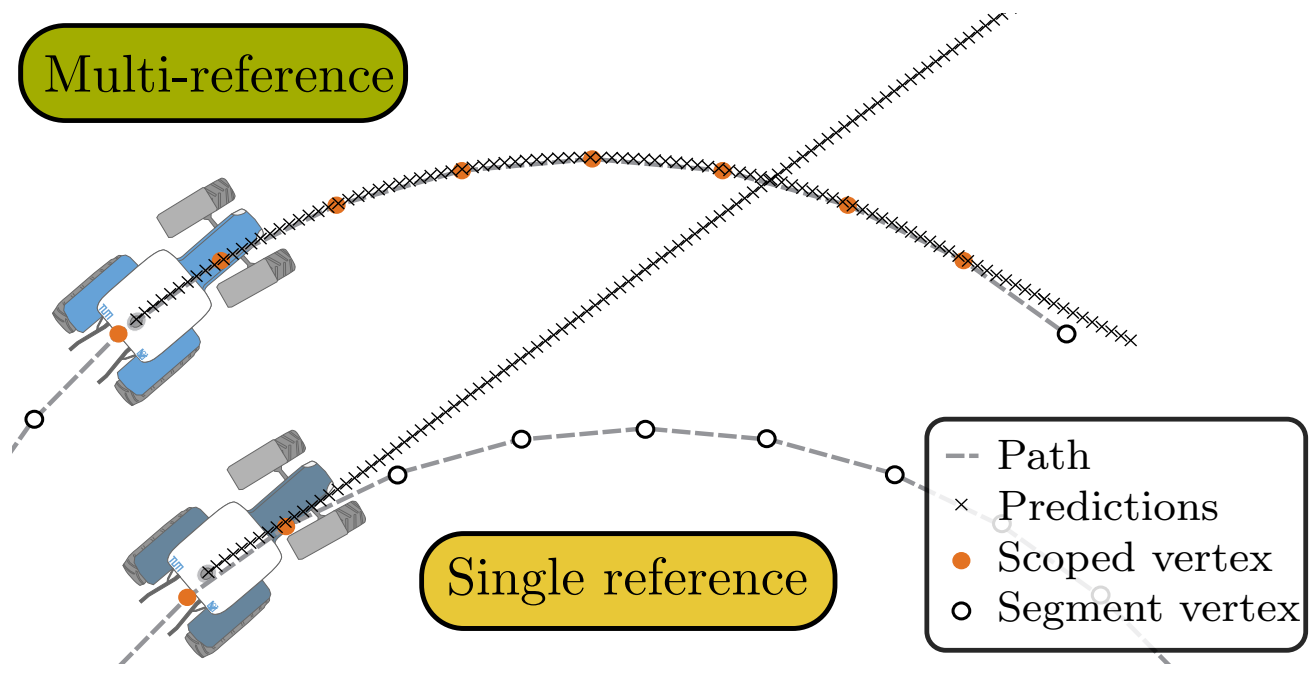

Fig. 1. Single-timestep comparison of a tractor-based guidance system using Nonlinear Model Predictive Control. Top: Controller predicting over multiple reference segments. Bottom: Controller restricted to a single reference segment.

Another approach, referred to as path tracking in this work, does not include an explicit time association and instead relies on a space-based trajectory reference. Its advantage is that the controller is not driven to catch up to a timed schedule once it falls behind, making it more suitable for general agricultural operations where exact timing is secondary. An example is presented in Lenain et al. (2005), where a MPC augments a nonlinear control law to compensate for overshoots caused by actuator delays. The underlying system model is formulated in curvilinear coordinates, directly coupling the reference path with the vehicle kinematics in the system equations.

An alternative formulation for a path tracking controller has been proposed by Backman et al. (2012), which keeps the kinematic system model unchanged, but includes the deviation from a linear reference path in the cost function. While this cost-based approach is robust to tune and straightforward to analyze, it is limited to a single linear reference segment, as illustrated in the lower part of fig. 1. Especially on curved path sections, this limitation can lead to a saw-tooth-shaped tracking error and prevents the controller from fully benefiting from longer prediction horizons, since the path is not correctly represented.

### *1.2 Scope of this Work*

This paper extends the Nonlinear Model Predictive Control path tracking controller of Backman et al. (2012) by enabling the tracking of multiple reference segments within the prediction horizon. The objective is to obtain the tracking behavior illustrated in the upper part of fig. 1, thereby improving the controller's predictive capability on curved paths. In addition, the approach is implemented within an experimental guidance system, including custom path segment handling and integration with the Tractor Implement Management steering interface.

## 2. METHODS

### *2.1 Nonlinear Model Predictive Control*

Following Diehl and Gros (2024), one of the main objectives of MPC is to achieve closed-loop optimal feedback control in real time by simplifying the underlying optimal control problem. In practice, the optimization is solved only on a finite prediction horizon using algorithms tailored for real-time execution. Once a solution is obtained, only part of the computed control sequence is applied to the system, typically the first control input.

During each control cycle, the optimization is solved with updated initial conditions, parameters or constraints derived from feedback of the real system. In this work, the MPC problem to be solved is nonlinear and formulated as follows:

$$\begin{aligned} \min_{\boldsymbol{x}(\cdot),\boldsymbol{u}(\cdot)} \quad & \int_0^{t_f} \boldsymbol{L}\big(\boldsymbol{x}(t),\boldsymbol{u}(t),\boldsymbol{p}(t)\big)\mathrm{d}t \\ & + \boldsymbol{E}\left(\boldsymbol{x}(t_f),\boldsymbol{p}(t_f)\right) \\ \text{subject to} \quad & \dot{\boldsymbol{x}}(t) = \boldsymbol{f}(\boldsymbol{x}(t),\boldsymbol{u}(t),\boldsymbol{p}(t)), \\ & \boldsymbol{x}(\boldsymbol{0}) = \boldsymbol{x_0}, \\ & \boldsymbol{x}_{\min} \le \boldsymbol{x}(t) \le \boldsymbol{x}_{\max}, \\ & \boldsymbol{u}_{\min} \le \boldsymbol{u}(t) \le \boldsymbol{u}_{\max} \quad . \end{aligned} \tag{1}$$

The cost function to be minimized consists of a Mayer term $\boldsymbol{E}$ and a Lagrange term $\boldsymbol{L}$, where the latter is formulated as a nonlinear least-squares expression:

$$\boldsymbol{L}\big(\boldsymbol{x}(t),\boldsymbol{u}(t),\boldsymbol{p}(t)\big) = \frac{1}{2}\Big\|\boldsymbol{y}\big(\boldsymbol{x}(t),\boldsymbol{u}(t),\boldsymbol{p}(t)\big) - \boldsymbol{y}_{\mathrm{ref}}\Big\|^2_{\boldsymbol{W}} . \tag{2}$$

In (2), the vector $\boldsymbol{y}$ has the combined dimensions of states and controls, but may otherwise contain arbitrary nonlinear functions of the states $\boldsymbol{x}$, control inputs $\boldsymbol{u}$, and parameters $\boldsymbol{p}$. The differences to the desired reference $\boldsymbol{y}_{\mathrm{ref}}$ is weighted by $\boldsymbol{W}$. The Mayer term $\boldsymbol{E}$ of (1) has a similar form as (2), but is not further detailed since it is not utilized in the proposed controller. The evolution of the states is introduced by the nonlinear system $\boldsymbol{f}$, starting from the initial states $\boldsymbol{x}_0$ and subjected to box constraints with upper ($_{\max}$) and lower ($_{\min}$) boundaries for both states and controls.

### *2.2 System Model*

The model utilized in (1) is a continuous time kinematic bicycle model, with an additional state for the steering angle derivative $\dot{\delta}^t$. The nonlinear system dynamics are given by

$$\boldsymbol{f}(\boldsymbol{x},\boldsymbol{u},\boldsymbol{p}) = \underbrace{\begin{bmatrix} x^{\mathrm{g}} \\ y^{\mathrm{g}} \\ \psi^{\mathrm{g}} \\ \delta^t \end{bmatrix}^{\cdot}}_{\dot{\boldsymbol{x}}} = \begin{bmatrix} v^{\mathrm{t}}\cos(\psi^{\mathrm{g}}) \\ v^{\mathrm{t}}\sin(\psi^{\mathrm{g}}) \\ v^{\mathrm{t}}\frac{\tan(\delta^t)}{l_{\mathrm{w}}} \\ \dot{\delta}^{\mathrm{t}} \end{bmatrix} , \quad u = \dot{\delta}^{\mathrm{t}}. \tag{3}$$

In (3), the reference frames of the variables are distinguished by indices, where the superscript g denotes the global and t the local tractor coordinate system. The coordinates $x^{\mathrm{g}}$ and $y^{\mathrm{g}}$, as well as the heading $\psi^{\mathrm{g}}$, are expressed in a North-East-Down tangent-plane frame. The speed $v^{\mathrm{t}}$ and steering angle $\delta^{\mathrm{t}}$ are defined in the tractor coordinate system, where the $x$-axis points forward along the wheelbase $l_w$ and the $y$-axis points to the right, as seen from the driver's seat.

The control variable $u$ used by the controller is the steering-angle derivative. This allows constraints to be imposed directly on the steering rate $\dot{\delta}^{\mathrm{t}}$. For this study, the velocity $v^{\mathrm{t}}$ is assumed to be manually controlled by the operator, which is common in tractor-guidance systems. It is therefore treated as a constant parameter that may change between controller iterations.

The kinematic representation is adopted due to the real-time requirements of the NMPC controller and the limited set of system parameters required, which are directly available via the ISO 11783 Controller Area Network (CAN) bus.

### *2.3 Cost function*

*Single-Segment Reference* The main cost function is based on the principle of calculating a cross-track distance relative to a linear line segment as introduced in Backman et al. (2012). By definition, the shortest pseudo-distance $\tilde{d}$ between a line and a point $P = (x, y)$ is where the connection vector is normal to the line:

$$\tilde{d} = \frac{(y_2 - y_1)\,x - (x_2 - x_1)\,y + x_2 y_1 - y_2 x_1}{\sqrt{(y_2 - y_1)^2 + (x_2 - x_1)^2}} . \tag{4}$$

Note that $\tilde{d}$ can be negative, which is achieved by removing the absolute value from the nominator in (4). This preserves the information about the side of the reference line on which the point $P$ lies, enabling the controller to determine the appropriate sign of the steering command.

For finite line segments, the cross-track metric is valid only if the origin of the connecting vector lies within the bounds of the reference segment. This condition can be verified by evaluating the progress

$$s = \frac{(x - x_1)(x_2 - x_1) + (y - y_1)(y_2 - y_1)}{(y_2 - y_1)^2 + (x_2 - x_1)^2} \; . \tag{5}$$

The progress is normalized by the length of the line segment, such that $0 \leq s \leq 1$ indicates that the point on the path remains within the segment.

*Multi-Segment Reference* To enhance the predictive capability of the controller, it is desirable to include multiple segments of the path into the cost function. A piecewise linear reference that depends on the system states is not well suited for Newton-type optimizers, because the resulting reference function is non-smooth at the segment boundaries. Therefore, a method for approximating the change between references is introduced based on sigmoid functions

$$\sigma(s, s_0) = \frac{1}{1 + \exp\bigl(-k\,(s - s_0)\bigr)} \; . \tag{6}$$

The slope of (6) can be controlled via $k$, where a larger value produces a steeper transition. Under nominal conditions, the transition is centered around zero, but can be shifted using $s_0$.

For the following, it is assumed that a reference path with $n_{\text{ref}} \geq 3$ points and $n_{\text{seg}} = n_{\text{ref}} - 1$ segments is provided.

The desired behavior of the distance cost term $c_{\tilde{d}}$ is to yield the cross-track metric for exactly one reference segment, namely the one closest to the predicted position. To achieve this, two intermediate terms are constructed to activate the cost terms based on the progress $s$ along the $i$th segment:

$$\boldsymbol{\alpha} = \begin{bmatrix} 1 \\ \alpha_1 \sigma(s_2, 0) \\ \vdots \\ \alpha_{n-1}\sigma(s_n, 0) \end{bmatrix} \quad , \quad \boldsymbol{\beta} = \begin{bmatrix} 1 \\ \beta_1 \sigma(s_1, 1) \\ \vdots \\ \beta_{n-1}\sigma(s_{n-1}, 1) \end{bmatrix} \; . \tag{7}$$

The combined distance cost is then

$$\begin{aligned} c_{\tilde{d}} = {} & \bigl[1 - \sigma(s_1, 1)\bigr]\tilde{d}_1 \\ & + \sum_{i=2}^{n_{\text{seg}}-1} \Bigl[\beta_i \alpha_i \bigl(1 - \sigma(s_i, 1)\bigr)\Bigr]\tilde{d}_i \\ & + \bigl[\beta_{n_{\text{seg}}} \alpha_{n_{\text{seg}}} \sigma(s_{n_{\text{seg}}}, 0)\bigr]\tilde{d}_{n_{\text{seg}}} \; , \end{aligned} \tag{8}$$

where the first and last selectors are not clamped by the segment boundaries. This allows the controller to maintain a viable reference even if the predicted position lies before the first or beyond the last segment.

The term $\alpha_i$ in (7) ensures that a new segment can only become active once the previous segment has been activated. The term $\beta_i$ becomes active once the previous segment has been completed. Both are necessary because the progress $s$ alone does not guarantee a unique activation.

For instance, multiple segments can have a progress $0 \leq s \leq 1$ during a curve, where the vehicle is on the concave side of the arc described by the path.

The remaining sigmoid function in (8) finally takes care of closing the associated window.

Assuming infinitely steep sigmoid functions, the proposed method ensures an exclusive activation of a segment-related distance cost. It does however not guarantee that at least one segment is active.

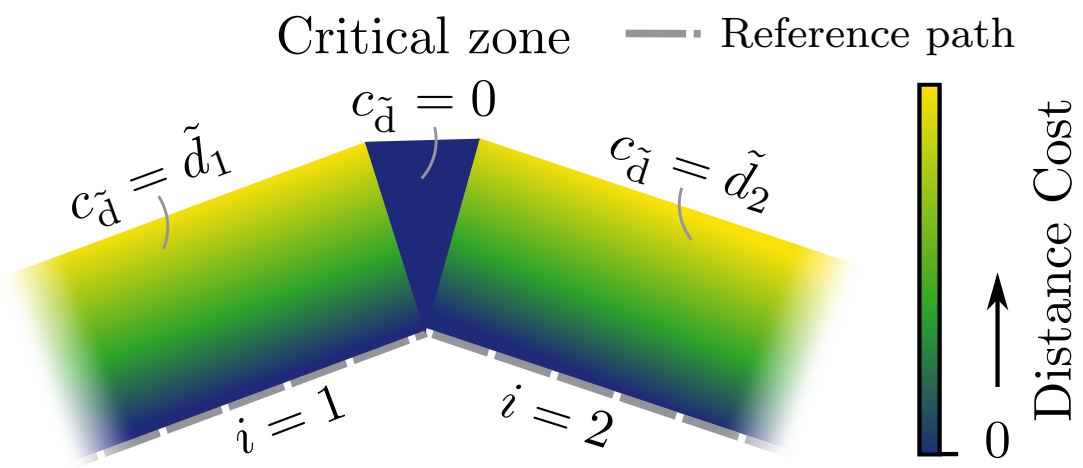

Fig. 2. Critical zone of the multi-reference distance cost in (8) for an outer curve, showing the region where the cross-track metric is undefined and results in zero cost between two adjacent reference segments.

Consider the case of the outer curve shown in fig. 2. Between two segments with opposing orientations, there exists a region where no valid cross-track metric is defined. According to (8), the distance-related cost becomes zero in this region, which may mislead the optimizer.

When the sigmoid functions are evaluated with a finite steepness $k$, the cost landscape is altered. Figure 3 shows the sampled distance-cost term $c_{\tilde{d}}$ for three different values of $k$ applied to the same reference path. A small value of $k$ pushes the critical regions away from the actual path, but at the same time, blending between segments distorts the underlying piecewise-linear reference. Increasing the steepness improves the approximation, but also moves the critical regions closer to the path.

For the case $k = 3$, the critical zones in the particular example remain sufficiently far outside the relevant tracking region, which typically extends only a small distance from the reference path. In contrast, $k = 20$ can easily lead to a diverging control behavior. In general, $k$ acts as a tuning parameter, dependent on both the segment length and the relative angle between consecutive segments. Both parameters can be controlled by resampling the reference segment and constraining the minimum turning radius.

*Combined Cost* In addition to the distance term $c_{\tilde{d}}$, a simple cost for the steering-angle derivative $\dot{\delta}^{\text{t}}$ is introduced, which multiplies the corresponding weight from $\boldsymbol{W}$ by the rate of change of the steering actuator. This term is intended to minimize the required actuator effort, providing a tuning parameter to trade off a smoother ride with less frequent control actions against faster error correction. To this end, the assembled cost vector to be used in (2) is

$$\boldsymbol{y} = \bigl[c_{\tilde{d}} \;\; 0 \;\; 0 \;\; 0 \;\; \dot{\delta}^{\text{t}}\bigr]^T \; . \tag{9}$$

### *2.4 Runtime Parameter Updating*

The cost function and dynamic system depend on parameters

$$\boldsymbol{p} = \bigl[\boldsymbol{x}^{\text{g}}_{\text{ref}} \;\; \boldsymbol{y}^{\text{g}}_{\text{ref}} \;\; v^{\text{t}} \;\; k \;\; l_{\text{w}}\bigr] \; , \tag{10}$$

which are included in the optimization problem but treated as constants during each iteration. Because these parameters determine the feasibility of the optimal control problem, special care must be taken to ensure they are provided consistently to the controller.

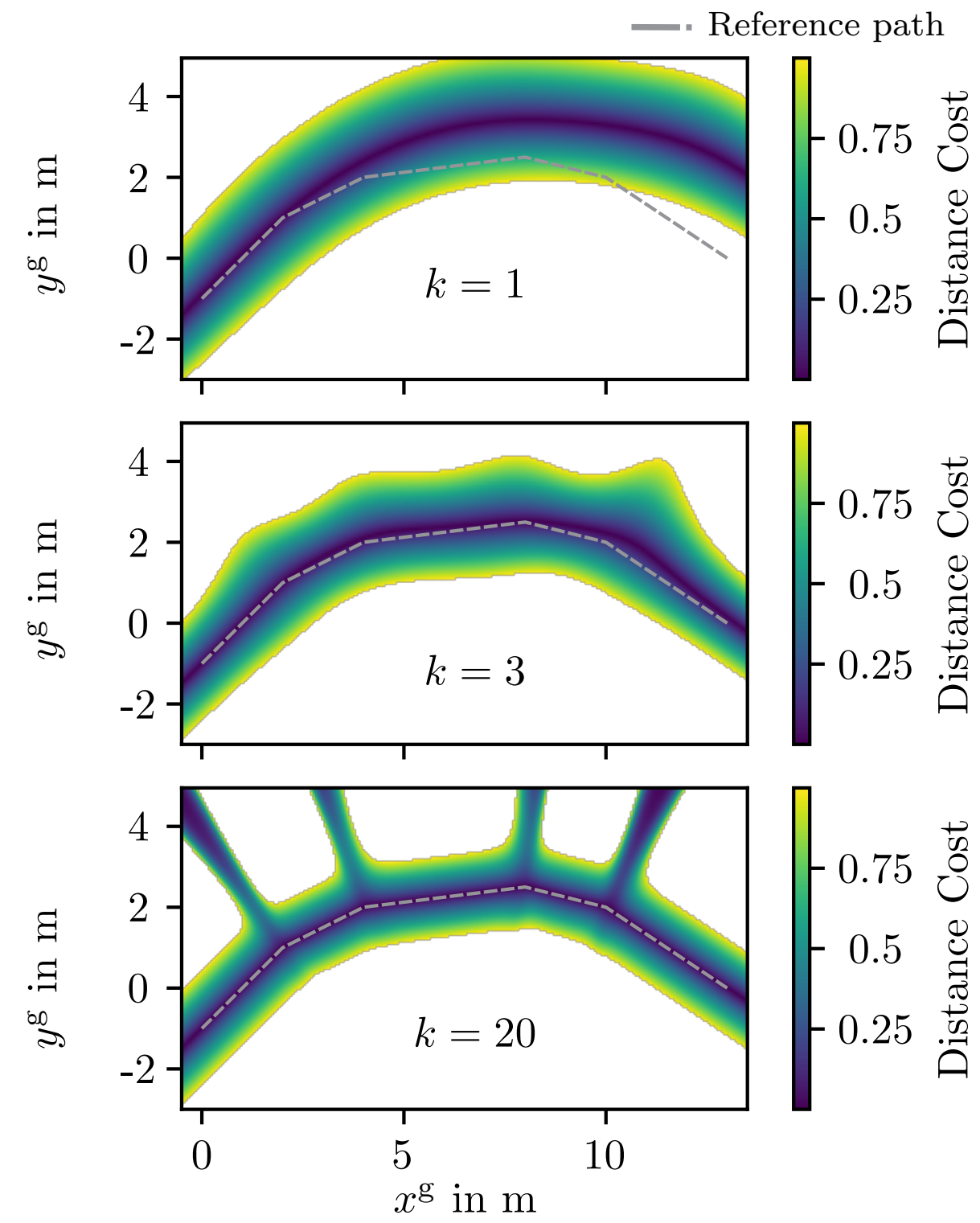

Fig. 3. Distance–cost evaluation for different values of the smoothing factor $k$ on the same reference path (gray). For improved visual appearance, only cost values below 1 are shown.

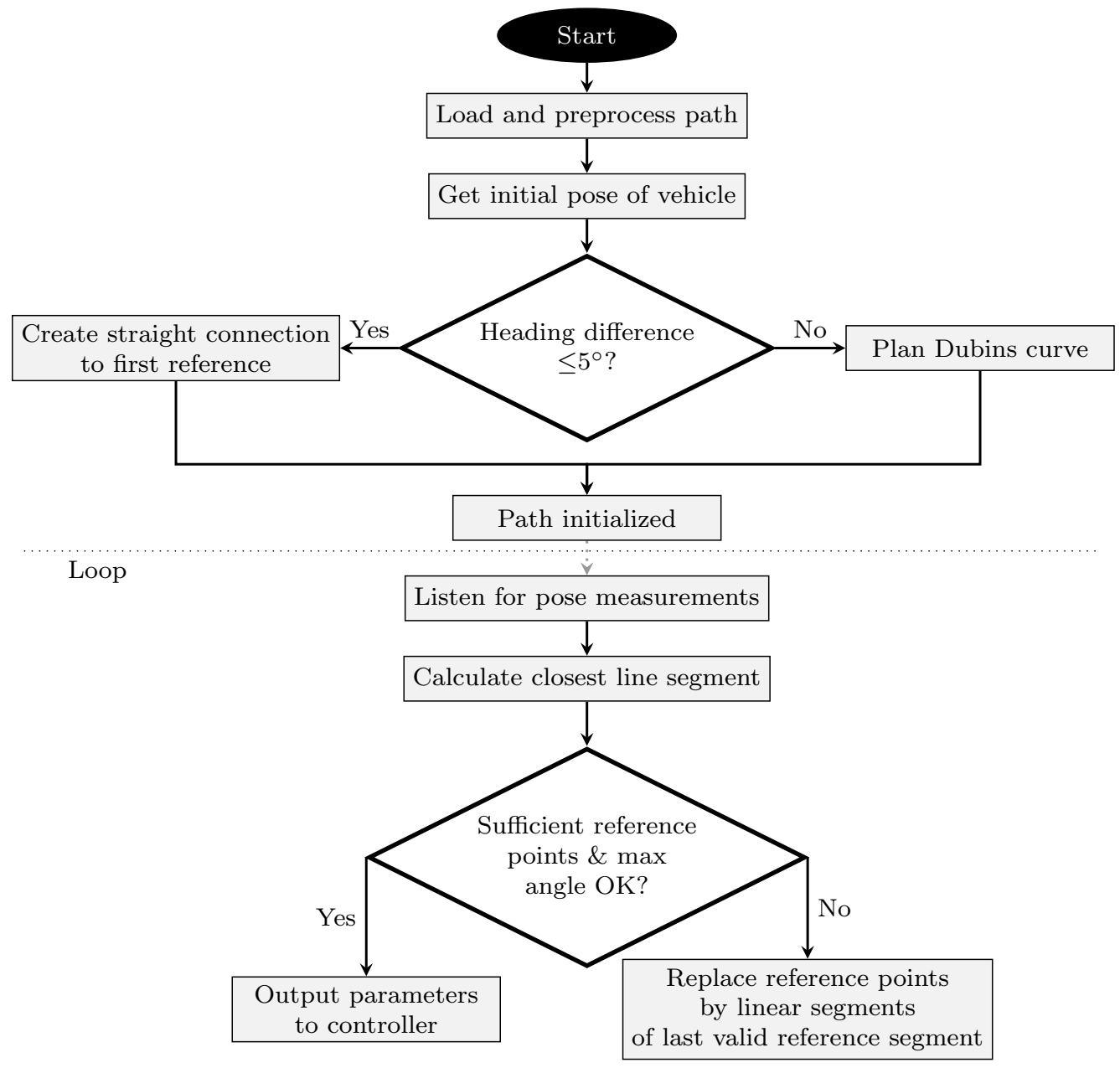

Fig. 4. Flowchart of the path handler for selecting valid parameters.

Whereas the sigmoid slope $k$ and the wheelbase $l_\mathrm{w}$ in (10) typically remain unchanged once tuning and calibration are complete, the tractor velocity must be updated based on the most recent measurement. The reference path, consisting of points $\boldsymbol{P}^g_\mathrm{ref} = (\boldsymbol{x}^\mathrm{g}_\mathrm{ref}, \boldsymbol{y}^\mathrm{g}_\mathrm{ref})$, is generally only a subset of a previously planned path. To initialize the path and consecutively select viable reference segments, a path handler is implemented, structured as shown in fig. 4.

After loading the reference path from memory, a preprocessing step ensures that all segments exceed a minimum length, thereby preventing path elements from being skipped during operation. Subsequently, a startup trajectory is generated to provide a feasible initial solution for the controller. Depending on the vehicle's orientation relative to the first segment of the main path, either a straight connection or a Dubins path, as introduced by Dubins (1957), is created. The maximum curvature of the startup path is limited to respect the steering constraints.

Once the path is initialized, the path handler is executed in every controller iteration using the current vehicle position and orientation. It maintains the current path index, representing the segment that was previously closest to the vehicle. This is determined using (4) and (5) together with routines designed to resolve ambiguous situations.
Only two adjacent segments are considered when updating the path index. This prevents the path from being skipped in overlapping regions but requires that each segment is reachable and has sufficient length.

Before passing the associated parameters to the controller, two special cases are handled. First, if the reference path does not provide a sufficient number of points, the last segment is linearly extended. Without this measure, (8) would produce `NaN` values during optimization, as missing parameters are initialized to zero by default. Second, the orientation difference between the first and subsequent segments is computed. If the angle exceeds 60°, the following segments are temporarily replaced with linear extensions of the last valid segment. This prevents the first segment from being reactivated once the vehicle has completed a turn.

### *2.5 Real-time implementation*

For the real-time implementation, the NMPC framework *acados* as introduced by Verschueren et al. (2021) is used. It offers a convenient interface for formulating optimal control problems with CasADi by Andersson et al. (2018). Interfaces are available for Matlab, C, and Python to define the problem structure and automatically generate efficient C code. In a rapid-prototyping approach, the initial controller development was performed within a Matlab simulation environment, where the generated controller was conveniently embedded via the MEX interface provided by *acados*. The finalized controller was then integrated into a C++ program, which also manages communication with the machine and handles the reference path.
The QP solver selected for this work is HPIPM (Frison and Diehl (2020)), as it efficiently handles the proposed cost formulation and solves the QP subproblems reliably. The prediction horizon is set to 100 steps with a fixed interval of 100 ms, corresponding to the maximum control frequency over an ISO 11783 CAN bus.

The tractor actuators are controlled via TIM as defined in AEF Guideline 023 RIG (2023). All sensor values and control commands are transmitted via a single ISO 11783 CAN bus.
The expected delays $t_{\mathrm{d}\delta}$ of the steering actuator are mitigated by propagating the system dynamics of (1)

using a Runge–Kutta integrator. During this prediction, previous control commands are applied according to the estimated delay, while the tractor speed is assumed to remain constant. The NMPC is then initialized with the predicted system state at the point in time at which the newly computed control input is expected to take effect. Note that prediction accuracy decreases in the presence of speed variations within the prediction horizon and system delay, which directly degrades tracking performance.

# 3. MATERIALS

The test tractor is a Fendt 314 Vario Gen 4 equipped with a TIM interface capable of controlling the steering system. Two GNSS antennas are mounted on the roof of the tractor, aligned with the vehicle's $x^{\mathrm{t}}$-axis with a baseline of 1.625 m. Both antennas are connected to a Septentrio mosaic-H GNSS receiver, which is integrated with a custom microcontroller featuring a CAN interface. Differential corrections are supplied to the receiver via a network connection to an on-site reference station, enabling fast carrier-ambiguity resolution and a reliable RTK fix. While the geodetic position of the main antenna has an estimated accuracy of 2 cm, the receiver reports the standard deviation of the heading angle between the antennas with 0.15°.

The most recent steering angle of the tractor is obtained from the TIM message set, whereas the speed is retrieved via the standardized ISO 11783-7 (2022) message. Custom routines for message handling, the TIM client, and inter-process communication have been implemented, but are not discussed further due to space constraints.

# 4. RESULTS

For evaluating the core concepts of the controller, a reference path with varying curvature has been created. The figure-eight shaped path has been tracked on a grassland field. The parameters, constraints and weights of the controller were set according to table 1.

Table 1. Parameters and weights as used in the field trial.

| Name | Variable | Value |
|---|---|---|
| Weights | $\boldsymbol{W}$ | $\mathrm{diag}\begin{bmatrix}3 & 0 & 0 & 0 & 0.01\end{bmatrix}$ |
| Sigmoid slope | $k$ | 3 |
| Wheelbase | $l_{\mathrm{w}}$ | 2.42 m |
| Actuator delay | $t_{\mathrm{d}\delta}$ | 300 ms |
| Speed | $v^{\mathrm{t}}$ | $1.2\,\mathrm{m\,s^{-1}}$ to $2.5\,\mathrm{m\,s^{-1}}$ |
| Steering bounds | $\delta^{\mathrm{t}}$ | $-30° < \delta^{\mathrm{t}} < 30°$ |
| Steering der. bounds | $\dot{\delta}^{\mathrm{t}}$ | $-40\,°\,\mathrm{s^{-1}} < \dot{\delta}^{\mathrm{t}} < 40\,°\,\mathrm{s^{-1}}$ |

## *4.1 Tracking Accuracy*

The main metric for evaluating the path tracking accuracy is the cross-track error of (4) and (5). In cases where no valid cross-track error is defined, the closest distance to the path nodes is used instead. The test path, including the automatically generated start and end segments, is shown in the upper part of fig. 5. Overall, the controller is able to track the curved reference path effectively. Upon closer inspection of the mid part in fig. 5, an oscillating cross-track error becomes visible, which results partially from the linear line segments and the smoothing residuals. In addition, there are various regions with an increased cross-track error, which reaches a maximum of −26.5 cm and 32.9 cm. On average, the cross-track error is 0.2 cm with a standard deviation of 8.4 cm. This aligns with the observed driving behavior: During turns, the controller tracks a slightly increased radius compared to the planned reference, which cancels out due to the symmetric path. The absolute value of the cross-track metric evaluates to 6.1 cm.

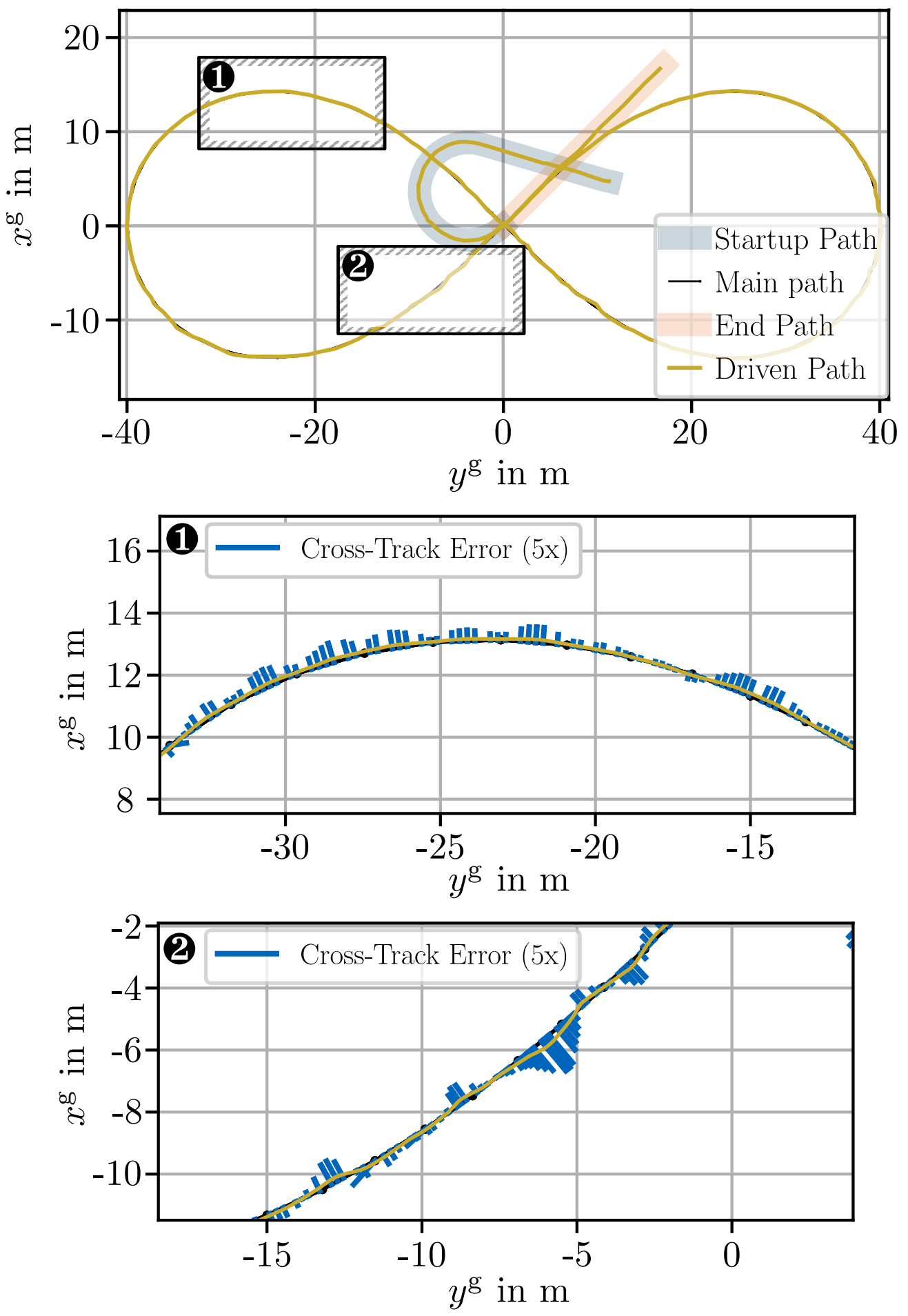

Fig. 5. Figure-eight reference path with the recorded tracking results and the automatically generated start and end segments (top). Bottom: Close-up views of the highlighted sections, showing the cross-track error amplified by a factor of five for visualization.

As illustrated in the lowest section of fig. 5, there are several occurrences of local oscillations observable. Since no Inertial Measurement Unit was available on the test-machine, the oscillations may stem from a missing projection of the GNSS measurements to the ground level. Especially when the vehicle is driving over a bump in the uneven terrain, a temporary roll angle can lead the controller to make an unnecessary course correction.

## *4.2 NMPC Solution Statistics*

The execution time of the *acados*-based NMPC implementation was recorded during the field trial. Each mea-surement begins when the control calculation is initiated and ends once the solver returns. The full set of timing measurements is shown in fig. 6, with individual samples color-coded according to the number of Sequential Quadratic Programming (SQP) iterations. On average, a solution was available after 3.45 ms, with a maximum of 9.0 ms. Throughout the entire test, the solver reached convergence, requiring a median of 2 and a maximum of 5 SQP iterations. The execution time thus remains well below the anticipated control cycle time of 100 ms, even with the long prediction horizon of 100 steps.

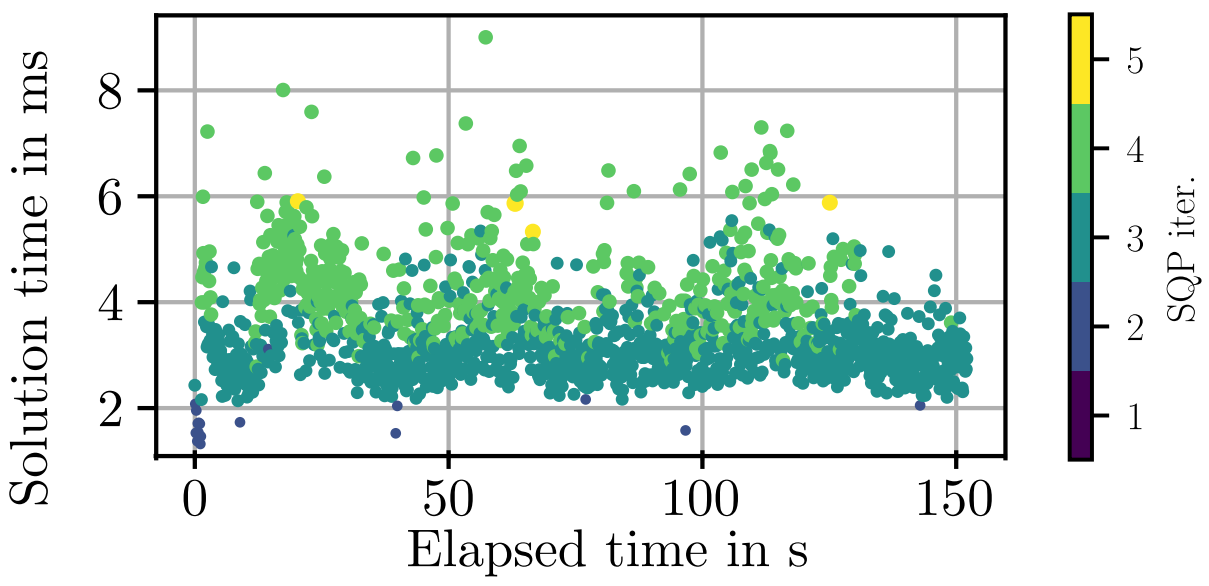

Fig. 6. Measured execution times of the proposed NMPC controller during the field trial. Time measurements are color-coded based on the number of SQP iterations required for convergence. All results were recorded on an AMD Ryzen 7530U.

## 5. CONCLUSION

This work has focused on developing a tractor-guidance system capable of incorporating multiple linear reference segments into the objective of an NMPC path tracking controller. With this addition, the control system can make use of longer prediction horizons compared to previous implementations. Alongside the controller, a method for selecting the required line segments from a predefined path was introduced. The controller was successfully tested during a field experiment using the tractor's TIM steering interface. Even though the evaluated setup does not match the accuracy or refinement of commercial integrated guidance systems, the real-time capability of the NMPC controller, despite the long prediction horizon and the extensive cost formulation, seems promising. In future work, applying the NMPC methodology to more challenging systems may enable more advanced use cases that are not viable with current guidance systems.

## ACKNOWLEDGEMENTS

The authors would like to thank AGCO GmbH for providing the tractor with the TIM interface. In addition, the authors appreciate the support from Michael Maier regarding the GNSS sensor and from Samuel Brodie for the assistance with the Tractor Implement Management client.

## DECLARATION OF GENERATIVE AI AND AI-ASSISTED TECHNOLOGIES IN THE WRITING PROCESS

During the preparation of this work the authors used various LLM tools in order to improve the spelling and gramatical correctness of the article.